\documentclass[10pt,twocolumn,letterpaper]{article}

\usepackage[pagenumbers]{cvpr}              
\usepackage{multirow}
\usepackage{float}
\usepackage{array}
\usepackage{tabu}
\usepackage[accsupp]{axessibility}

\usepackage[dvipsnames]{xcolor}

\definecolor{cvprblue}{rgb}{0.21,0.49,0.74}
\usepackage[pagebackref,breaklinks,colorlinks,citecolor=cvprblue]{hyperref}

\def\paperID{69} 
\def\confName{EarthVision}
\def\confYear{2024}

\title{Contrastive Pretraining for Visual Concept Explanations \\ of Socioeconomic Outcomes}

\author{Ivica Obadic$^{1,2}$
\quad
Alex Levering$^{3}$
\quad
Lars Pennig$^{1}$
\quad
Dario Oliveira$^{4}$
\quad
Diego Marcos$^{5}$
\quad
Xiaoxiang Zhu$^{1,2}$
\\
{\small $^1$Technical University of Munich}  \quad {\small $^2$Munich Center for Machine Learning} \quad {\small $^3$Vrije Universiteit Amsterdam} \\ {\small $^4$Getulio Vargas Foundation} \quad {\small $^5$ Inria, University of Montpellier}
}

\begin{document}
\maketitle

\begin{abstract}
Predicting socioeconomic indicators from satellite imagery with deep learning has become an increasingly popular research direction. Post-hoc concept-based explanations can be an important step towards broader adoption of these models in policy-making as they enable the interpretation of socioeconomic outcomes based on visual concepts that are intuitive to humans. 
In this paper, we study the interplay between representation learning using an additional task-specific contrastive loss and post-hoc concept explainability for socioeconomic studies.
Our results on two different geographical locations and tasks indicate that the task-specific pretraining imposes a continuous ordering of the latent space embeddings according to the socioeconomic outcomes. This improves the model's interpretability as it enables the latent space of the model to associate concepts encoding typical urban and natural area patterns with continuous intervals of socioeconomic outcomes. 
Further, we illustrate how analyzing the model's conceptual sensitivity for the intervals of socioeconomic outcomes can shed light on new insights for urban studies.
\end{abstract}

\section{Introduction}\label{chapter:introduction}

\begin{figure*}[t]
    \centering
        \includegraphics[width=\textwidth]{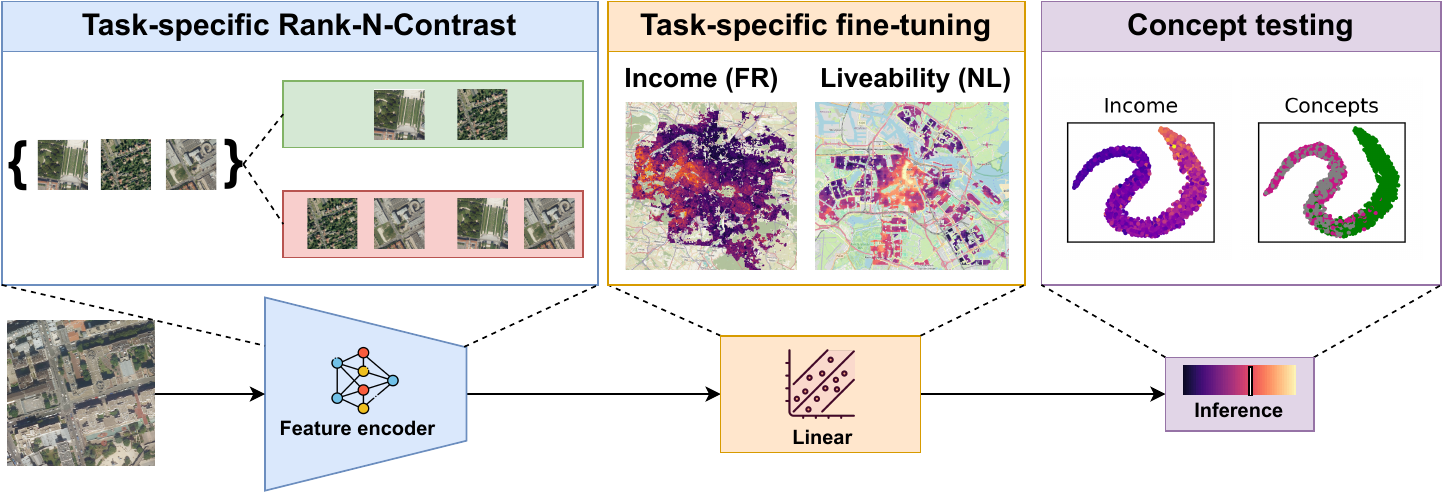}
        \caption{\textbf{Flowchart of the pipeline presented in this research.} For both tasks, we use Rank-n Contrast to pre-train the feature encoder to produce embeddings that strongly relate to socioeconomic task scores. Secondly, we freeze the encoder weights and probe a linear layer to regress the task-specific score. Lastly, we use TCAV to study the relation of various urban concepts to socioeconomic scores.}
        \label{fig:method}
\end{figure*}

Monitoring socioeconomic outcomes is crucial for effective policy-making and progress toward the Sustainable Development Goals (SDGs). An emerging approach in recent years is to predict these indicators by training deep learning models on satellite imagery \cite{burke2021using}. For example, such approaches are applied for estimating the economic well-being in Africa \cite{yeh2020using, ni2020investigation}, the income in France \cite{abitbol2020interpretable}, the urban vitality in Italy \cite{scepanovic2021jane}, the landscape scenicness in the UK \cite{levering2021relation}, or the livability of cities in the Netherlands \cite{levering2023predicting}. 
However, these approaches are rarely used in practice by decision-makers \cite{burke2021using}, with one of the main obstacles being the opaque nature of the deep learning models, whose inner workings and mechanisms lack human interpretability \cite{hall2022review}. 

The interpretability challenges in the proposed deep learning models for socioeconomic indicator estimation have usually been addressed with saliency map methods like Grad-CAM \cite{abitbol2020interpretable} or concept bottleneck models \cite{ scepanovic2021jane, levering2021relation, levering2023predicting}.  While saliency maps identify relevant regions in the image by highlighting raw pixels, they do not provide additional insights into the high-level semantics the model extracted for its inference \cite{achtibat2023attribution}. Furthermore, they are of questionable reliability as they can be insensitive to the changes in the model weights or data distribution \cite{Adebayo}.
A promising alternative to the saliency maps methods are the concept-based explanations that relate a target variable to a set of explanatory concepts. Compared to the saliency maps, they more closely resemble human reasoning and thus are more intuitive \cite{User_Study}. Moreover, concept-based explanations are a natural fit for socioeconomic studies, as concepts are commonly used as indicator variables \cite{lbm_report}. These approaches for remote sensing have only been attempted through bottleneck models, which relate intermediate concepts to an output variable in an end-to-end trained manner \cite{scepanovic2021jane, levering2021relation, levering2023predicting}.  However, such methods also have several downsides. They require that a dataset of annotated concepts specific to the study region is available during training to condition the bottleneck and assume a strong relation between the concepts and the target variables. The post-hoc concept-based explanations can overcome these issues as they relate the outcomes to a set of human interpretable concepts after the model training phase \cite{Zhou, pmlr-v80-kim18d, ghorbani2019towards}. These methods identify the concepts by looking for a direction in the latent space of the trained model towards which the concept examples are found. The concepts are represented with coherent images sharing similar visual characteristics; thus, the resulting explanations are intuitively understandable \cite{pmlr-v80-kim18d, ghorbani2019towards}. Post-hoc concept explanation methods have already been successfully applied in other sensitive domains like medicine \cite{pmlr-v80-kim18d, graziani2020concept}. These approaches can also be beneficial for remote sensing studies, as, unlike concept bottleneck models, they do not require an additional set of concept labels specific to the geographical area of the study region. 

In this work, we introduce a new expandability pipeline (visualized in Figure~\ref{fig:method}) that uses post-hoc concept explanations to reveal how deep learning models predict socioeconomic outcomes in terms of easy-to-understand concepts that represent typical urban and natural area patterns. It consists of three main steps: 1) a task-specific contrastive pre-training of a deep learning encoder to continuously order the latent space embeddings based on their socioeconomic outcome, 2) probing a linear layer on the top of the encoder to predict socioeconomic outcome, and 3) concept testing with the Testing with Concept Activation Vectors (TCAV) method \cite{pmlr-v80-kim18d}. We evaluate our pipeline on two tasks for estimating socioeconomic indicators from remote sensing imagery, namely income prediction in France \cite{abitbol2020interpretable}, and liveability estimation in the Netherlands \cite{levering2023predicting}. In summary, our work evaluates for the first time visual post-hoc concept-based explanations in remote sensing socioeconomic studies with the following main contributions: 
\begin{itemize}
    \item Continuously ordering the latent space embeddings based on their socioeconomic outcomes yields comparable or improved prediction performance to a baseline supervised learning.
    \item Such latent space improves the interpretability of the deep learning model as the concepts cluster according to continuous intervals of socioeconomic outcomes. 
    \item Finally, this enables studying the sensitivities of socioeconomic outcomes to different urban concepts.
\end{itemize}

\section{Related Work}
\label{chapter:relatedwork}
Although essential for their wider adoption in practice, the interpretability of the deep learning models for socioeconomic indicator estimation is rarely tackled \cite{hall2022review}.  The approaches in related studies mainly propose using saliency maps \cite{abitbol2020interpretable} or concept bottleneck models \cite{scepanovic2021jane, levering2021relation, levering2023predicting}.
Concretely, \citeauthor{abitbol2020interpretable} train an EfficientNet deep learning model on aerial imagery to predict household income in five French cities \cite{abitbol2020interpretable}. They used the Grad-CAM saliency map method to infer the relevant regions for the model predictions. The identified regions are matched with the urban class polygons to reveal the relationship between urban classes and the income prediction of the deep learning model. Their results show that residential areas can both activate in case of high or low socioeconomic status depending on the city and area, whereas the presence of infrastructure such as motor- or railways is consistently related to a lower household income.
When it comes to concept bottleneck models, they are used by \citeauthor{scepanovic2021jane} to predict the urban vitality in six Italian cities from Sentinel-2 imagery \cite{scepanovic2021jane}. In the first step, the authors use a VGG-16 network and a convolutional autoencoder to predict six different land use and building block characteristics from image patches. These features are then used as input to the linear model that estimates urban vitality. The authors discover, that intersection density, small parks, anisotropicity, and land use mix positively influence vitality, while the variables block size and building height negatively impact the vitality. 
Further, \citeauthor{levering2021relation} train an end-to-end convolutional neural network with a semantic bottleneck to predict the landscape scenicness in the UK based on Sentinel-2 imagery \cite{levering2021relation}.
The interpretable concepts in the semantic bottleneck layer encode the region's land cover distribution. This architecture enables an explanation of the predicted scenicness score in terms of the land cover characteristics of the region. The authors find, that scenic areas are often associated with mountainous, coastal regions, while urban areas and the presence of human influence are negatively related to the landscape scenicness. 
The semantic bottleneck models are also used in 
\cite{levering2023predicting} to predict the livability in the Dutch cities using aerial imagery. Again, the authors include a semantic bottleneck layer, which predicts five different intermediate domain scores. Their results indicate, that some intermediate domains such as buildings or physical environments are much better suited to be predicted from aerial imagery than others like safety or amenities. 

While the above-mentioned approaches shed light on the relevant factors for socioeconomic indicator prediction, they rely on the availability of labeled features specific to the study region. Hence, the limited data availability in remote sensing represents an obstacle to applying these approaches to different geographical areas. Further, saliency maps are known to produce local, low-level feature explanations with limited faithfulness \cite{Adebayo}. 
In this work, we aim to overcome these limitations by explaining the relevant factors for monitoring socioeconomic indicators through high-level concepts using the TCAV method that can be applied to various geographies without requiring concept labels specific to the study region.
\section{Methods}\label{chapter:methods}
In this section, we detail our explainability pipeline presented in Figure~\ref{fig:method} for post-hoc concept-based explanations of socioeconomic outcomes. These methods usually identify concepts as directions in the latent space of an already trained model. This makes their interpretation challenging, as the post-hoc analysis might yield redundant and ambiguous concepts \cite{Concept_whitening}. To overcome this, we rely on a contrastive pretraining technique to enforce continuous ordering of the latent space embeddings based on the target outcome. Such a representation can improve the interpretability of the tested concepts, as it forces them to align along intervals of the target outcome,  In summary, the presented approach consists of three steps. Firstly, we train the feature encoder using Rank-N-Contrast (RNC) loss, which optimizes the feature extractor to produce embeddings that are continuously ordered according to the target outcome. Next, we probe a linear layer on the frozen embedding space to regress an output score. Lastly, we analyze the learned embedding through Testing using Concept Activation Vectors (TCAV), a post-hoc concept-based interpretability method.

\subsection{Rank-N-Contrast}
Rank-N-Contrast (RNC) is a contrastive feature learning method for regression that uses the proximity in label values to form positive and negative pairs within a given batch. In this study we utilize it to pre-train the feature extractor to provide feature embeddings that are aligned with the target regression scores, thereby guiding the expressivity of features in the latent space towards only being relevant for the regression task. Specifically, RNC \textit{"ranks the samples according to their target distances, and then contrasts them against each other based on their relative rankings"} \cite{zha2024rank}. Consider an anchor image $\mathbf{v}_i$ within a batch of images. For this anchor and a pairing candidate $\mathbf{v}_j$, it samples pairs which satisfy the condition $S_{i,j} := \{ v_k \mid k \neq i, d(\tilde{y}_i, \tilde{y}_k) \geq d(\tilde{y}_i, \tilde{y}_j) \}$, where $d(\cdot, \cdot)$ is a label distance metric, while $\tilde{y}_i$, $\tilde{y}_j$ and $\tilde{y}_k$ are the labels of the $i$-th, $j$-th and $k$-th image in the batch, respectively. Subsequently, the feature vectors of all sample pairs are compared using the average negative log-likelihood over all pairings in the batch. By iterating over each sample as an anchor, we can calculate the following loss for the entire batch:
\begin{equation}
    \mathcal{L}_{RNC} = \frac{1}{2N}  \sum\limits_{i=1}^{2N} \frac{1}{2N-1} \sum\limits_{j=1, j\neq i}^{2N} -log(\sigma (v_i, v_j, S_{i,j}),
\end{equation}
where the Softmax function $\sigma$ is given as follows:
\begin{equation}
    \sigma (v_i, v_j,  S_{i,j}) = - \log \frac{\exp(\text{sim}(v_i, v_j)/\tau)}{\sum_{k\in  S_{i,j}}\exp(\text{sim}(v_i, v_k)/\tau)},
\end{equation}
where $\tau$ is a temperature parameter, and $\text{sim}(\cdot, \cdot)$ measures the similarity between the feature embeddings for a pair of images in the batch. Minimizing this loss enforces the feature encoder to produce similar embeddings for images with similar label values, thus ordering the image embeddings in the latent space according to their labels. After this pre-training step, a simple linear layer can be added to the model and trained on the frozen feature embeddings to regress the actual regression labels.

\subsection{Testing with Concept Activation Vectors}
\label{sec:methods_tcav}
We utilize \emph{Testing with Concept Activation Vectors} (TCAV) \cite{pmlr-v80-kim18d} as an approach for testing the relatedness of concepts to socioeconomic outcomes. TCAV allows uncovering meaningful directions in the latent space of a model based on a previously defined concept dataset and can measure the model's sensitivity to these concepts. Each concept is represented by a set of images sharing similar visual characteristics. These images are usually derived from external probe dataset \cite{ramaswamy2023overlooked} or automatically learned from the dataset \cite{ghorbani2019towards}. The workflow of TCAV can be divided into two main steps: 1) learning Concept Activation Vectors (CAVs) in the latent space of the trained model, and 2) testing the sensitivity of a concept to the model predictions.

\subsubsection{Learning Concept Activation Vectors}
A concept activation vector (CAV) represents a direction in the latent space where the examples of a concept are located. To learn a CAV in a hidden layer of the model, first, the datasets of images for the target concept, and for a random concept are fed into the trained deep learning model, and their activations from the target hidden layer are stored. These activations are used as input to a linear classifier that learns to discriminate the activations of images representing the concept against the activations of the images of the random concept. The normal vector of the hyperplane is interpreted as the CAV of the respective concept. CAVs can be trained in hidden layers of various depths of the model. This approach relies on the assumption of the linear separability of concepts in the latent space which is supported by several works \cite{linear1, linear2} which have shown that linear feature disentanglement can often be observed in the hidden layers.

\subsubsection{Concept sensitivity testing}
\label{sec:tcav_sensitivity_testing}
To evaluate the influence of a concept on the model outcomes, the conceptual sensitivity $S_{k,l}$ of a CAV $v_{l}$ in layer $l$ to the model output $k$ is determined. It is computed as the directional derivative of the output logit $h_{l,k}$ to $v_{l}$ with the following equation:
\begin{equation}
\label{eq:tcav_sensitivity}
    S_{k,l} = lim_{\epsilon \xrightarrow{} 0 } \frac{h_{l,k}(f_{l}(x)+\epsilon v_{l})-h_{l,k}(f_{l}(x))}{\epsilon}
    = \nabla h_{l,k} * v_{l}
\end{equation}
where $f_{l}(x)$ are the activations for the input $x$ at the layer $l$ of the deep neural network.
Thus, the sensitivity score can be related to change in the model prediction when an instance is perturbed in the direction of a concept. Next, the individual sensitivities are then aggregated to form the overall TCAV score. 
It is defined as the percentage of images that show a positive sensitivity to the CAV. 

\begin{equation}
    TCAV_{Q} = \frac{|x \in X_{k} : S_{k,l}(x) > 0|}{|X_{k}|}
\end{equation}

One caveat when computing the sensitivity scores is the calculation of the gradients. They capture first-order derivatives and thus may lead the attribution map to focus on irrelevant features \cite{Integrated_gradients}.
To circumvent this issue, we used attribution maps generated via \emph{Integrated Gradients} (IG) that can reduce the problematic locality of plain gradients. They are computed by aggregating the gradients of the model predictions for instances that lie along a straight line on the path from baseline image $x'$ to the input image $x$ with the following equation:
\begin{equation}
    \text{IG}(x) = (x-x') \int_{0}^{1}\frac{\partial h_{l,k}(f_{l}(x'+\alpha (x-x')))}{\partial x} d\alpha
\end{equation}
For images, the baseline is usually set to a black image, which marks the absence of image features.




\section{Experimental Setup}
\label{chapter:experimental_setup}
\subsection{Reference data}
We conduct our experiments on two tasks posing the problem of estimating socioeconomic indicators from aerial imagery. Firstly, we use a dataset that consists of aerial images paired with a winsorized average household income estimated from the tax return for 5 major cities in France in 2015 \cite{abitbol2020interpretable}. The images have a resolution of 20 cm and are of size 1000 x 1000. In line with the authors' workflow of training individual models for cities, we trained one model for Paris, which resulted in 18 446 pairs of aerial imagery and income labels. Similar to \cite{abitbol2020interpretable}, we first partition the instances according to the 5 quantiles of their income distribution. Subsequently, we perform stratified random sampling of 64\% of the data into training, and 16\% and 20\% in validation and test sets, respectively. For our second task, we use a dataset of liveability reference scores over the Netherlands, which combines the livability values derived from the Leefbaarometer project \cite{lbm_report} with aerial images made publicly available by the Dutch government. The resulting dataset covers 13 Dutch cities and includes 51.781 labeled image patches at a 1-meter resolution and a size of 500 x 500. We use the geographically stratified train/val/test splits made available by the authors \cite{levering2023predicting}. 

\subsection{Concept data}
We rely on land-cover classes to define visual concepts representing natural and urban areas that can be used to explain the model workings in socioeconomic studies. Concretely, we subset land cover annotations from the FLAIR dataset \cite{garioud2024flair}. It contains aerial imagery covering France at a spatial resolution of 20cm/pixel, as well as land cover class masks. In total, the labels differentiate between 19 different land cover classes, and in our experiments, we extract the following 7 visual concepts: \emph{Water, Natural Vegetation, Agriculture, Impervious Surface, Sparse Residential, Medium Residential}, and \emph{Dense Residential}. The first four concepts are simple visual primitives encoding mostly color and texture features. They are extracted from patches having $>90\%$ coverage of the following four subsets of land cover classes: \emph{Water, Natural Vegetation, Agriculture} and \emph{Impervious Surface}, respectively. The other three concepts represent more complex visual primitives that were extracted by aggregating land cover patches into three Local Climate Zone \cite{LCZs} classes based on the amount of built-up land visible in each image: \emph{Dense Residential ($>90\%$ Buildings), Medium Residential (40 to 60\% Buildings, 40-60\% Vegetation or Agriculture)}, and \emph{Sparse Residential (10-30\% Buildings, 70-90\% Vegetation or Agriculture)}. Visual examples of our classes can be found in the Appendix, in section~\ref{sec:classdefappendix}. While we use a set of pre-defined concepts in this research, it should be noted that concepts in remote sensing have stricter conditions on when they may be used. We discuss these conditions in more detail in section~\ref{sec: rsconcepts}.

\subsection{Inference and Model Training}
We predict the socioeconomic outcomes using a Resnet-50 encoder \cite{he2016deep} for feature extraction followed by a linear layer that regresses the socioeconomic outcome. To train the model, we followed a similar training procedure as in \cite{zha2024rank}: First, we optimize the encoder for 400 iterations to minimize the RNC loss (the hyperparameter details are provided in Section \ref{section:training_procedure} of the Appendix). Next, we freeze the weights of the feature encoder and train a single linear layer that uses the encoder output to predict the socioeconomic outcomes. The weights of the linear layer were optimized for 100 epochs with the $\mathcal{L}_1$ loss and we selected the model yielding the best $R^2$ score on the validation set. Our code is provided at {\small \url{github.com/IvicaObadic/rnc-4-visual-concept-explanations}}.
\section{Results}
\label{chapter:results}

\subsection{Prediction of Socioeconomic Indicators }
Table~\ref{tab:contrastive_pretraining_results} shows the effect of Rank-N-Contrast pretraining on the prediction results. It illustrates that the task-specific pretraining leads to a better fit for the household income dataset than the baseline model trained with an $L_1$ loss. 
Concretely, the $R^2$ improves for the contrastive pre-trained encoder by around 0.1 for both evaluation sets, validation, and test. Similarly, Kendall's $\tau$ correlation improves by around 0.08 on both sets.
Regarding the liveability dataset, the contrastive pretraining procedure yields results comparable results to the encoder directly trained for predicting livability scores with the $L_1$ loss. 

\begin{table}[h]
\caption{
\textbf{Impact of pretraining with Rank-N-Contrast (RNC) loss on prediction results. It results in a better prediction performance for the income predictions and comparable results for liveability.}
\label{tab:contrastive_pretraining_results}}
\centering
\scriptsize
\begin{tabular}{llllll}
\toprule
              &  & \multicolumn{2}{c}{\textbf{val set}} & \multicolumn{2}{c}{\textbf{test set}} \\ 
\textbf{dataset}  & \textbf{objective}  & \textbf{$R^2$} & \textbf{$\tau$}       & \textbf{$R^2$}       & \textbf{$\tau$}  
\\ \midrule

\multirow{2}{*}{\textbf{Income}} &  $\mathcal{L}_1$ loss  & 0.53             & 0.54    & 0.55         & 0.56                \\
& RNC ($\mathcal{L}_1$)  & 0.62             & 0.61    & 0.65         & 0.64                \\
\midrule

\multirow{2}{*}{\textbf{Liveability}} &  $\mathcal{L}_1$ loss  & 0.7             & 0.64    & 0.49         & 0.53                \\
& RNC ($\mathcal{L}_1$)  & 0.68             & 0.66    & 0.49         & 0.53 \\
\bottomrule
\end{tabular}
\end{table}

\begin{figure*}[t]
    \centering    
        {\includegraphics[width=\textwidth]{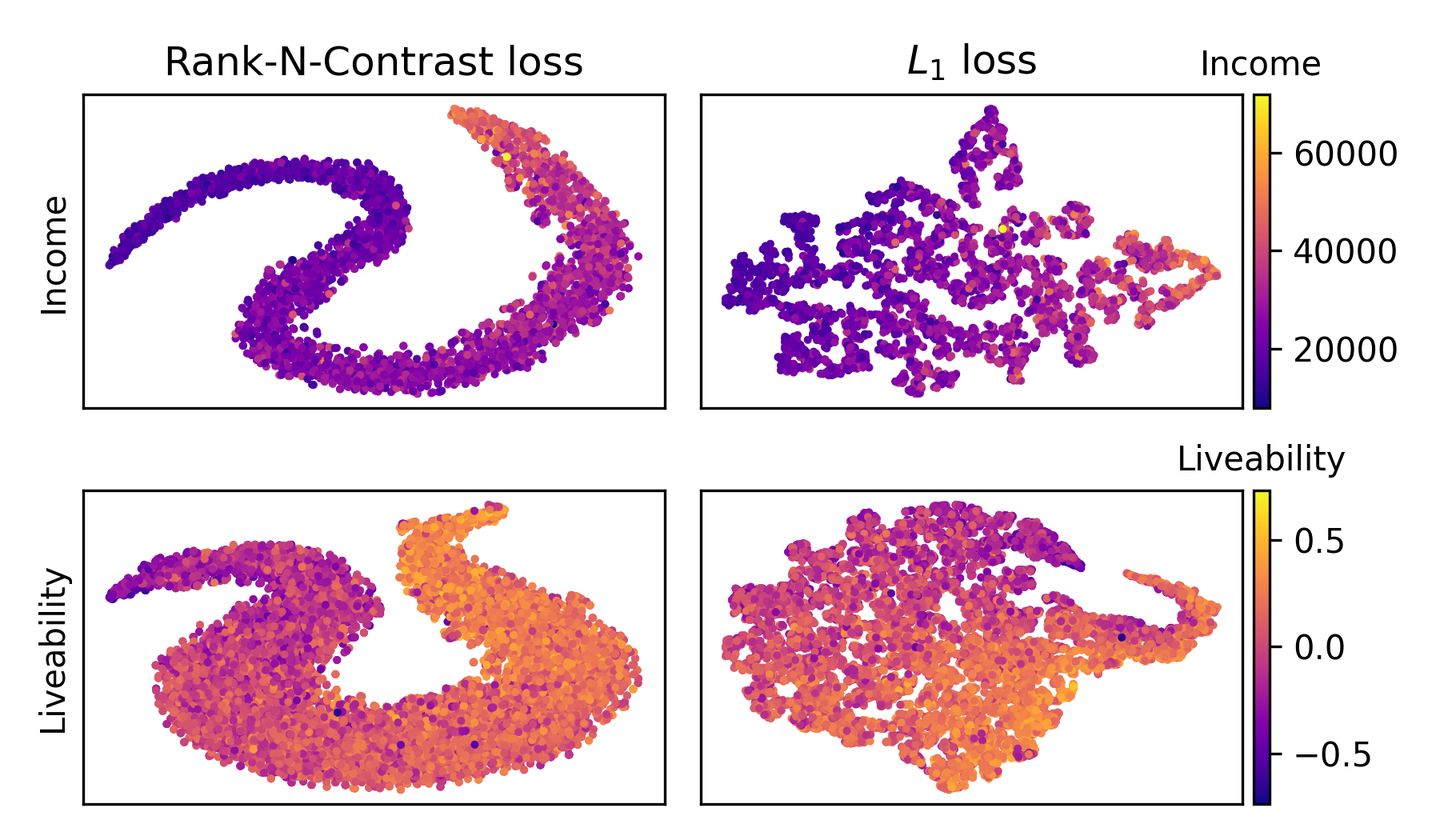}\label{fig:tsne_all_models}}
        \caption{\textbf{Household income instance activations in the average pooling layer visualized with t-SNE.} By pre-training with Rank-n Contrast, the latent space can be ordered according to the task regression values, rather than the visual features of images. This results in an embedding space that aligns with the socioeconomic outcomes, and therefore is better suited for interpretability.} 
        \label{fig:household_income_latent_space}
\end{figure*}

To better understand the effect of task-specific pretraining on the latent space, in Figure~\ref{fig:household_income_latent_space} we use the t-distributed Stochastic Neighbor Embedding (t-SNE) method \cite{van2008visualizing} to visualize the instance activations after the average pooling layer of the ResNet-50 encoder. 
The plots in the first column illustrate that the contrastive pre-trained encoder increasingly orders the instances in the latent space in an approximately 1D manifold according to their target label for both datasets. In contrast, the plots visualized in the second column depict that the models optimized to directly predict the regression score seem to result in a more scattered embedding space that fails to capture the continuous nature of the data, as there are multiple clusters of instances sharing similar target values located in various subspaces of the latent space. Finally, we note that for the task-specific pre-trained encoder, the liveability values typically display higher variance in their score compared to the income values for a local area in the embedding space. This can be attributed to the difference in complexity between these two tasks. The liveability reference data is based on 100 factors spanning 5 domains, from which liveability scores are derived \cite{levering2023predicting}. As it is a composite score, the ranking difference between pairs will depend on more complex patterns, which might explain the comparable prediction results in Table \ref{tab:contrastive_pretraining_results} of the task-specific pretraining with the supervised trained encoder on the liveability task.

\subsection{Concept Explanations}
\label{sec:concept_explanations}
\begin{figure*}[t]
    \centering    
    {\includegraphics[width=0.49\textwidth]{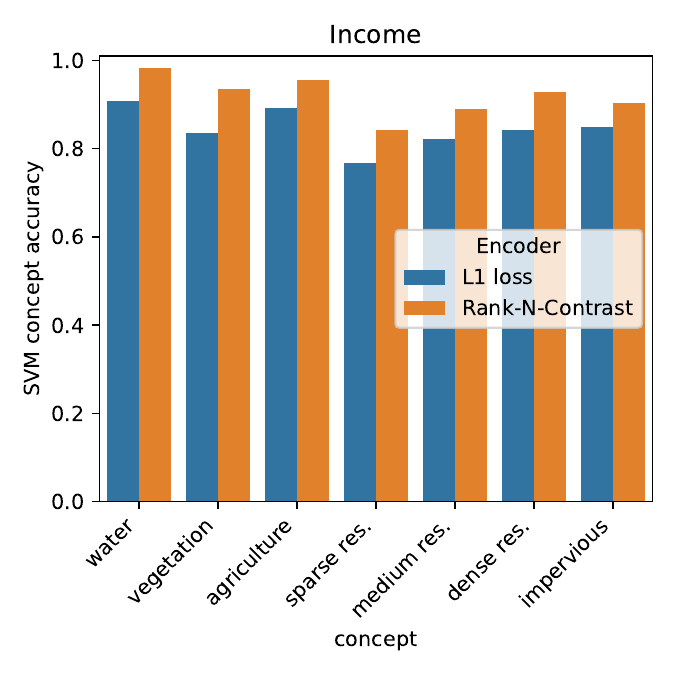} \label{fig:household_income_accuracy}}
    {\includegraphics[width=0.49\textwidth]{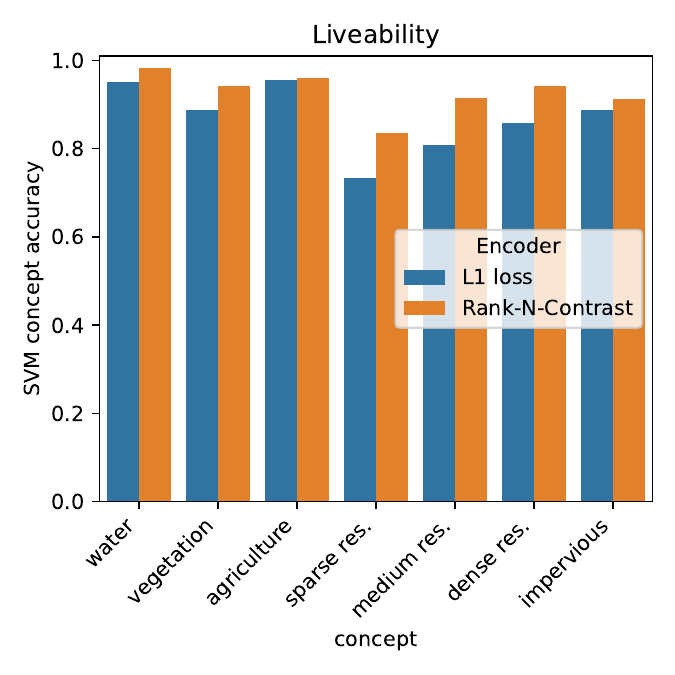}\label{fig:liveability_concept_layer_accuracy}}
    \caption{\textbf{Concept accuracy in the average pooling layer of the Resnet-50 encoder for the household income (left) and livability (right) datasets.} The contrastive pretraining improves the linear separability of the concepts in the latent space.} 
    \label{fig:concept_layer_accuracy}
\end{figure*}

\paragraph{Learning Concept Activation Vectors}
As described in Section~\ref{sec:methods_tcav}, an essential step in TCAV is learning a concept activation vector (CAV) that encodes the direction in the latent space where the examples of a concept are located.
In our study, we learn the CAVs per layer in the latent space of the ResNet-50 encoder by training a one-vs-all linear SVM classifier to discriminate the examples of the concept from randomly sampled 500 examples of all other concepts based on their layer activations. 
Figure~\ref{fig:concept_layer_accuracy} shows that while the concepts are already mostly linearly separable in the average pooling layer of the baseline encoder trained with the $L_1$ loss, enforcing a continuously ordered latent space with the Rank-N-Contrast pre-trained encoder further improves the concept discrimination. Notably, it enables consistently better linear separation for all concepts in the income task while for liveability the strongest improvements are observed in discriminating the sparse residential, medium residential, and dense residential concepts.

\begin{figure*}[t]
    \centering
        \includegraphics[width=\textwidth]{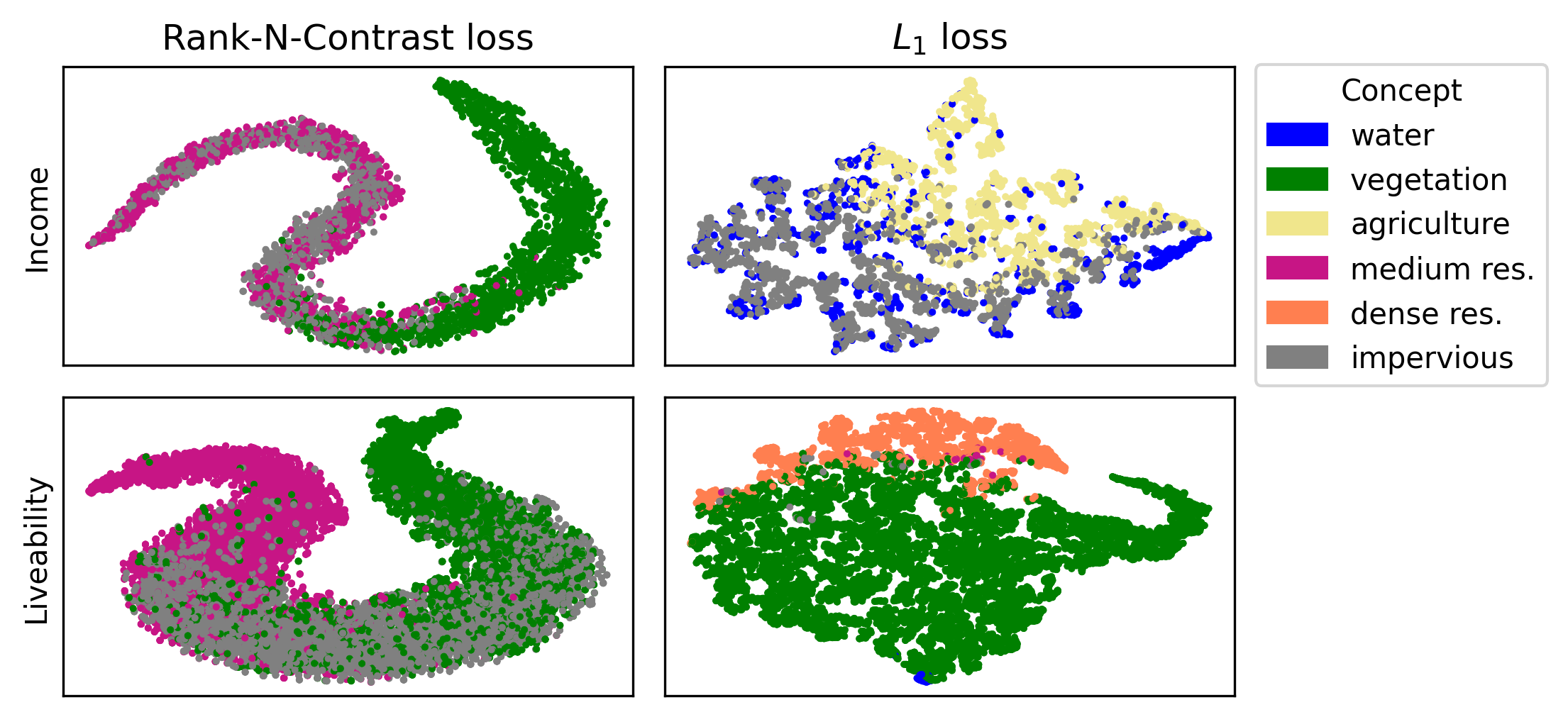}
        \caption{\textbf{Instance to concept alignment as seen in the average pooling layer of each model.} The instances are colored according to the concept with the highest cosine similarity (after normalizing the similarities per concept with the $L_2$ norm). The most similar concept to an instance can be interpreted as the concept that is most closely aligned to the instance's socioeconomic outcome.}
    \label{fig:concept_latent_space_allignment}
\end{figure*}

\paragraph{Alligning Concepts to Socioeconomic Outcomes}
Not imposing any constraints on the latent space during model training is challenging for post-hoc concept interpretability as the resulting model might yield ambiguous CAVs that are hard to interpret \cite{Concept_whitening}. 
As shown in the previous section, the Rank-N-Contrast pretraining procedure ensures that the embeddings lie on a manifold continuously ordered according to the target value. This impacts the learned CAVs as the concept examples are also projected along this manifold. Further, the high linear separability of the concepts on this manifold shown in Figure~\ref{fig:concept_layer_accuracy} suggests that the CAVs point to distinctive ranges of socioeconomic outcomes.  
We verify this by performing a similar analysis to \cite{pmlr-v80-kim18d} that identifies the most similar concept to each instance in the socioeconomic datasets based on the cosine similarity of the instance embeddings and the CAVs.
The graphs on the left-hand side of Figure~\ref{fig:concept_latent_space_allignment} reveal that in the contrastive pre-trained encoder, the concepts are clustered on typically continuous intervals of the target values. 
In detail, the left plot in the first row shows that lower to medium-income areas are associated with medium residential areas and impervious surfaces. Subsequently, the areas with higher-income typically relate to vegetation. Similarly, medium residential areas and vegetation also characterize the lower and the upper intervals of liveability, respectively. In contrast to the income task, the impervious surfaces occupy a larger portion of the liveability range, usually describing areas between the first and the third quartiles of liveability.
In summary,  these results show that the Rank-N-contrast pretraining procedure improves the interpretability of the concept explanations as it enables one to intuitively associate the concepts to a continuous interval of values for the socioeconomic indicators. 
Conversely, the plots on the right-hand side of Figure~\ref{fig:concept_latent_space_allignment} visualize that for both datasets, the models trained with the $L_1$ loss offer a limited understanding of how the concepts relate to the socioeconomic indicators as they typically do not form coherent clusters in the latent space. 

\begin{figure*}[t]
    \centering    
    {\includegraphics[width=\textwidth]{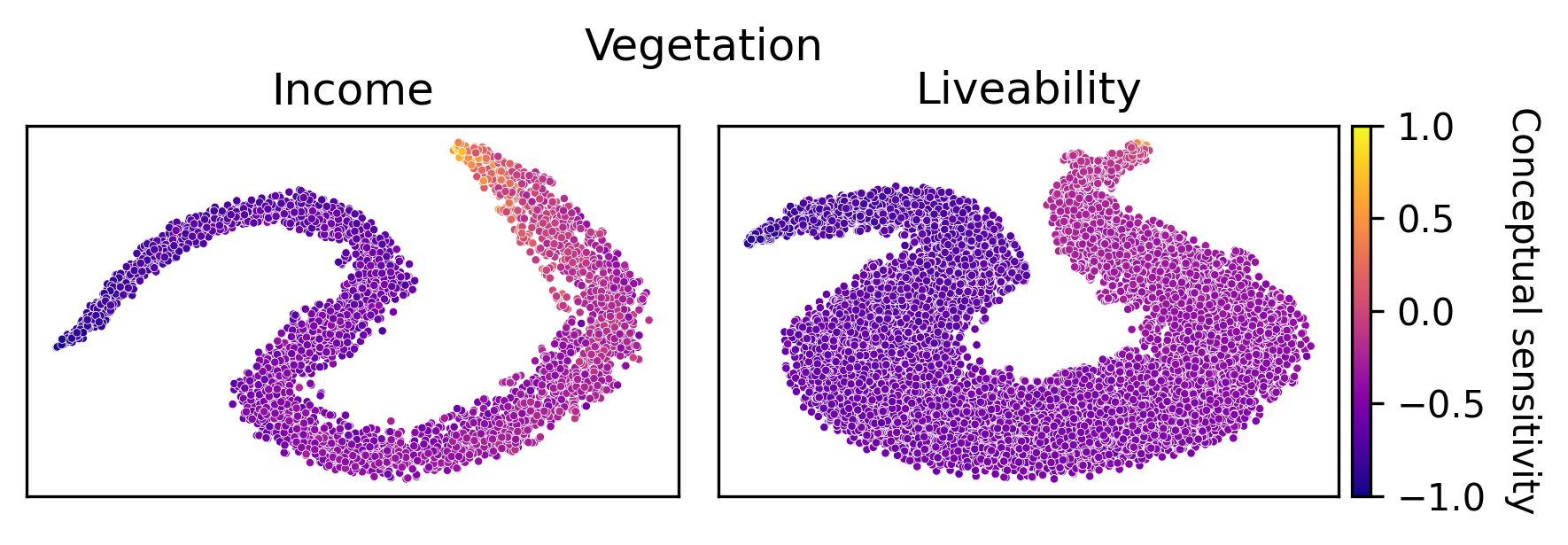}}
    \caption{\textbf{The TCAV sensitivity of the vegetation concept for the income (left) and liveability (right) datasets.} The magnitude values are normalized in the range [-1, 1] by applying separate min-max normalization to the negative and to the positive TCAV values, respectively. For income, the sensitivity to vegetation is highest among the high-income areas. In other words, adding vegetation to higher-income areas increases the perceived income of the neighborhood by the model. For liveability, we observe a similar effect, as the strongest increase in perceived liveability by the model in highly liveable areas can be achieved by increasing their amount of natural vegetation.} 
    \label{fig:tcav_values_vegetation}
\end{figure*}

\paragraph{Insights into Urban Planning with TCAV}
Measuring the sensitivity of the concept to the model predictions with Equation~(\ref{eq:tcav_sensitivity}) of Section~\ref{sec:tcav_sensitivity_testing} enables us to understand how the prediction changes when the visual patterns of the instances are perturbed in the direction of the concept. This unveils whether the concepts contribute positively or negatively to the perceived socioeconomic outcome by the model. Furthermore, the concept clusters resulting from Rank-N-Contrast pretraining explain these correlations of the various urban and natural area concepts with intervals of the socioeconomic outcomes. This is exemplified in Figure~\ref{fig:tcav_values_vegetation} which shows the sensitivities of the instances to the \emph{vegetation} concept. The sensitivities show the rural-urban gradient in income and liveability as the areas on the lower ends of both socioeconomic outcomes do not benefit much from adding vegetation, but its contribution and importance increase as income and liveability rise. Neighborhoods in more urbanized areas often have better amenities and higher wage expectations. However, access to nature is often lacking in these areas, which explains the positive contribution on the most positive ends of both socioeconomic outcome distributions. The sensitivity plots of other concepts shown in Section~\ref{sec:sensitivities} of the Appendix illustrate the diverging trends that emerge for the concepts among socioeconomic outcomes. While income shows mostly clear linear trends, liveability displays a more complicated, non-linear relationship with the concepts. This may be explained by the fact that the liveability score is a complex composite score spanning many dimensions, whereas income reflects a single socio-economic outcome. 


\subsection{Discussion - Defining Remote Sensing Concepts}
\label{sec: rsconcepts}
The use of concepts as a means of interpretability in natural images is well-studied, with many datasets available that contain concept classes. However, given the more complex and variable nature of remote sensing imagery, the use of concepts and their comparison between studies will strongly depend on definitions, datasets, and sensor resolution used. Firstly, as the resolution of an image decreases, individual objects lose their shape, and only geographical features, which are “an abstraction of a real-world phenomenon” \cite{iso_iso19101_2014}, remain. As a result, the definition of a geographical concept varies with the spatial scale that it is observed on \cite{woodcock_factor_1987}. Concept-centric approaches for remote sensing therefore need to contend with variable concept definitions, making large-scale remote sensing concept datasets harder to manage. Beyond concept definition uncertainties, common geographical products such as land cover maps may contain classes that are not visible with optical sensors \cite{congalton_global_2014}.
As Arvor et al \citep{arvor_ontologies_2019} note, geographic concepts suffer from vagueness in their definitions. Often, concepts are defined using qualitative metrics. For instance, “high NDVI” may define a forest. However, values ascribed to qualitative descriptors such as "high" may differ between studies. A second-factor affecting uniform concept definitions is determined by the ambiguity of a concept, where it has “no crisp boundaries and may shift its meaning over time as new technologies appear, as people develop new habits, and in general, as the social and physical context of the term evolves” \cite{wieringa_design_2002}. Finally, the cultural context of a term makes defining geographical concepts difficult, as these concepts may readily vary between cultures \cite{mark_toward_1993}. As concept-based explanations continue to mature, such considerations must be taken into account.

\section{Conclusion}\label{chapter:conclusion}
In this paper, we presented a pipeline that enables intuitive concept explanations of the deep learning models for socioeconomic outcome prediction. It consists of task-specific contrastive pretraining with the Rank-N-Contrast method and the TCAV approach for post-hoc concept testing. Our results show that the use of Rank-N-Contrast projects latent space embeddings in a continuously ordered manifold based on socioeconomic outcomes. This resulted in improved model performance for income prediction in Paris and yielded comparable results for liveability predictions in the Netherlands. The learned representation also contributed to intuitive concept explanations as we have shown that the concept activation vectors form clusters along continuous intervals of socioeconomic outcomes. For example, our analysis revealed that medium residential areas and impervious surfaces are associated with low to medium values of income and liveability.
In contrast, the areas that lie in the upper socioeconomic outcome intervals display a strong association with the vegetation concept. Further, ordering the concepts in the latent space is also beneficial for understanding their sensitivity to the intervals of socioeconomic outcomes. We used this property to demonstrate the rural-urban gradient for the vegetation concept and the diverging trends of the other concepts among the socioeconomic outcomes. Such findings open up the possibility of using our method to gain new insights into urban studies. Finally, we showed that our approach can be applied in different geographical locations as unlike concept bottleneck models, it does not require a set of concept labels specific to the location of the target study available during training.

\paragraph{Acknowledgments} This work is jointly supported by the Munich Center for Machine Learning and by the German Federal Ministry of Education and Research (BMBF) in the framework of the international future AI lab ``AI4EO -- Artificial Intelligence for Earth Observation: Reasoning, Uncertainties, Ethics and Beyond'' (grant number: 01DD20001). We thank Dino Ienco, Nassim Ait Ali Braham, and Andrés Camero for the fruitful discussions.

{
    \small
    \bibliographystyle{ieeenat_fullname}
    \bibliography{main}
}
\clearpage
\setcounter{page}{1}
\maketitlesupplementary

\section{Training Procedure}
\label{section:training_procedure}
For model training, we performed image transformations in the following order:
\begin{enumerate}
    \item Resizing the images to a size of 500 x 500
    \item Random Augmentation \cite{cubuk2020randaugment}
    \item Random Erasure \cite{zhong2020random}
    \item Min-max image normalization
\end{enumerate}
For the task-specific pretraining with the Rank-N-Contrast approach, we used the default hyperparameters suggested by the authors \cite{zha2024rank}. In detail, we set the temperature $\tau$ to 2, use negative $L_2$ norm as feature similarity metric, and $L_1$ distance as label distance. 
Further, this encoder was trained with the Adam optimizer and cosine rate annealing. Finally, we used the Adam optimizer with an exponential learning rate scheduler for the supervised training of the standard Resnet-50 encoder (as also implemented in \cite{levering2023predicting}) and the linear layer that probes the output of the task-specific pre-trained encoder.

\section{Concept Definitions}
Table \ref{tab:my_label} provides a detailed overview of the used land cover percentages for each of the concepts and visualizes examples of concept images.
\label{sec:classdefappendix}
\begin{table*}[!h]
    \centering
    \begin{tabular}{|m{30mm}|m{30mm}|m{15mm}|m{31mm}|}
    \hline
    \textbf{Concept} & \textbf{FLAIR Class} &  \textbf{Percent} &\textbf{Examples} \\
     \hline
       Water &   Water &   90-100 & \vspace{2mm}\includegraphics[width=15mm, height=15mm]{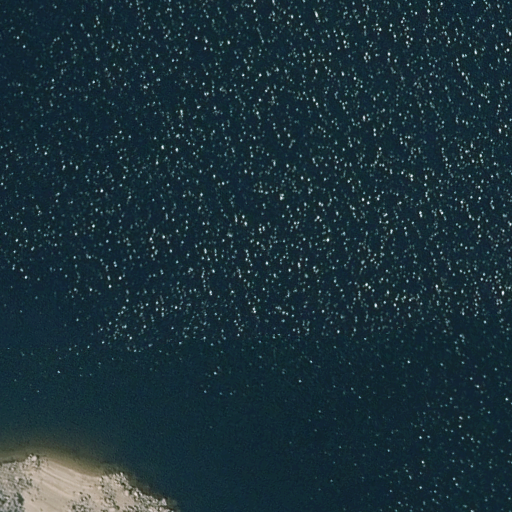} \includegraphics[width=15mm, height=15mm]{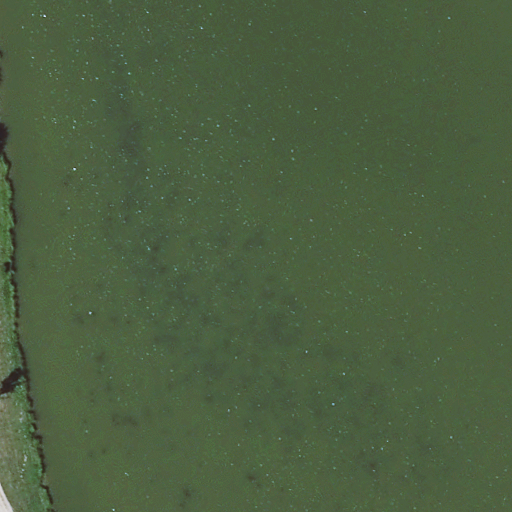} \\
     \hline
        Vegetation &Vegetation&90-100 & 
     \vspace{2mm}\includegraphics[width=15mm, height=15mm]{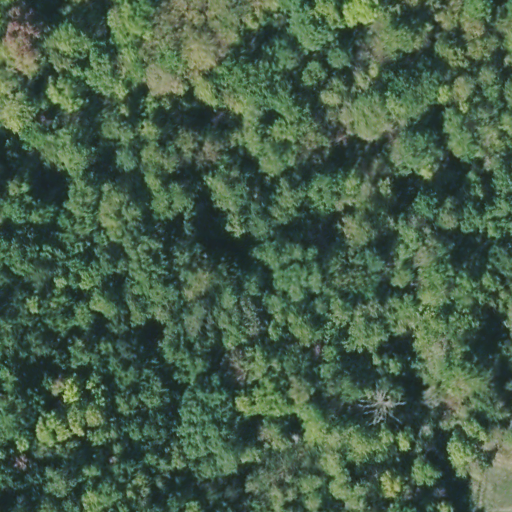} \includegraphics[width=15mm, height=15mm]{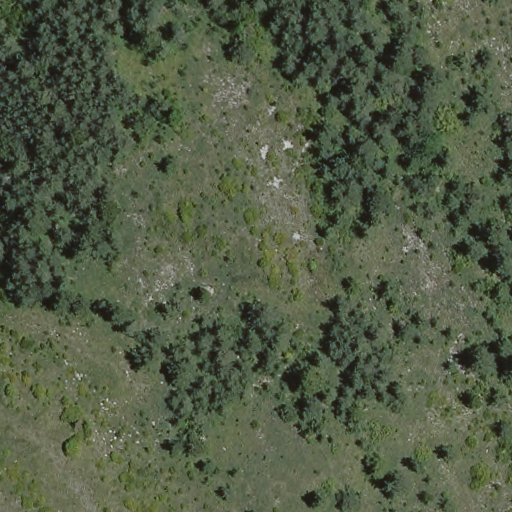}\\
     \hline
       Agriculture&Agriculture&90-100 &\vspace{2mm}\includegraphics[width=15mm, height=15mm]{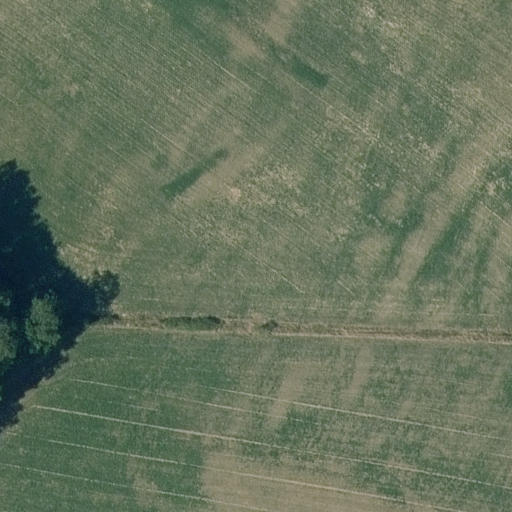} \includegraphics[width=15mm, height=15mm]{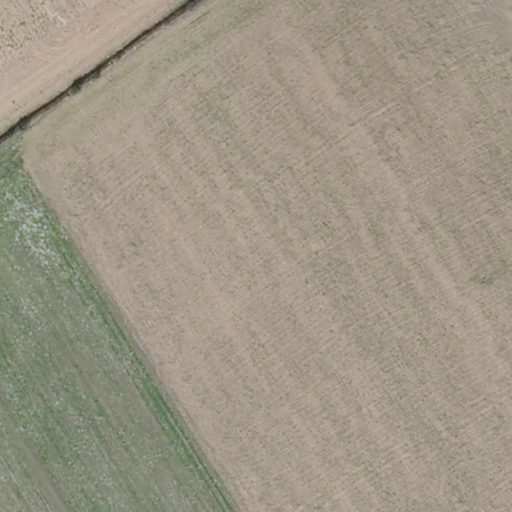}\\
     \hline
       Impervious \newline Surface&Impervious \newline Surface&90-100&\vspace{3mm}\includegraphics[width=15mm, height=15mm]{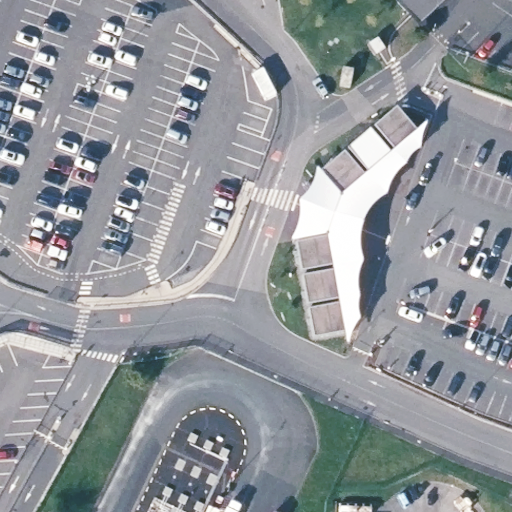} \includegraphics[width=15mm, height=15mm]{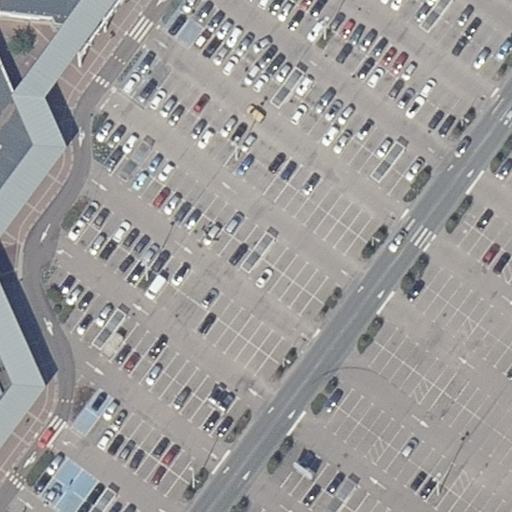} \\
     \hline
       Dense Residential&Buildings&90-100&\vspace{2mm}\includegraphics[width=15mm, height=15mm]{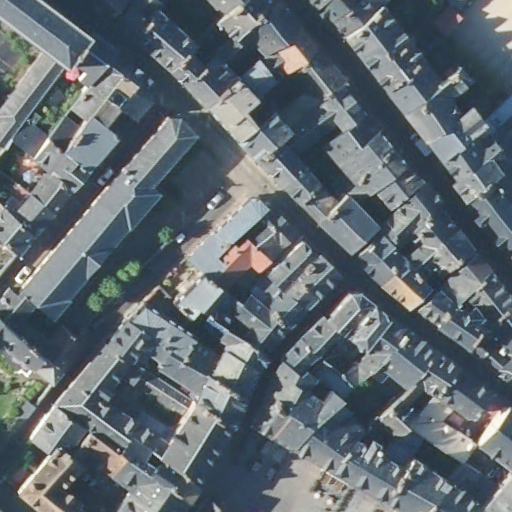} \includegraphics[width=15mm, height=15mm]{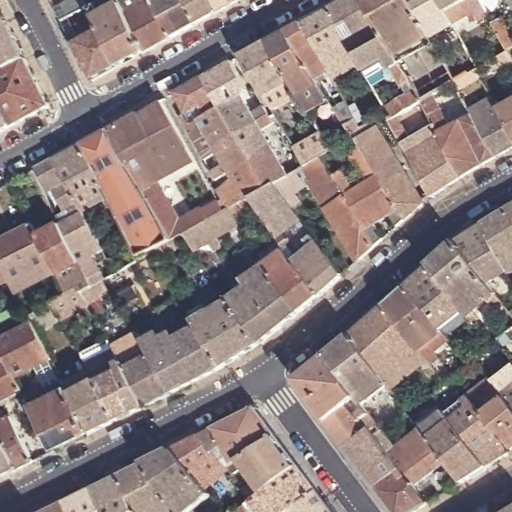} \\
     \hline
       \multirow{2}{30mm}{\newline Medium Residential}    & Buildings &  \vspace{3mm}40-60 & \vspace{4mm}\multirow{2}{35mm}{\includegraphics[width=15mm, height=15mm]{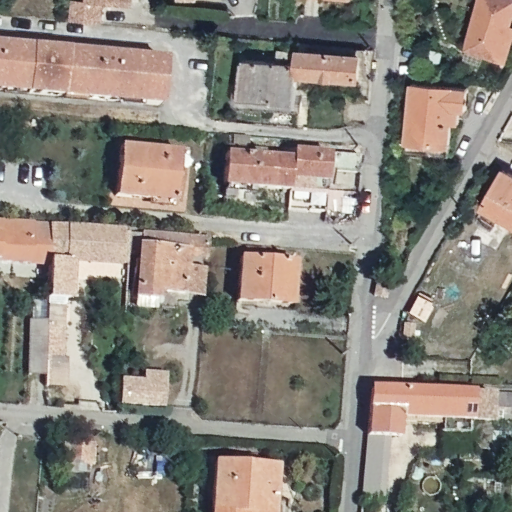} \includegraphics[width=15mm, height=15mm]{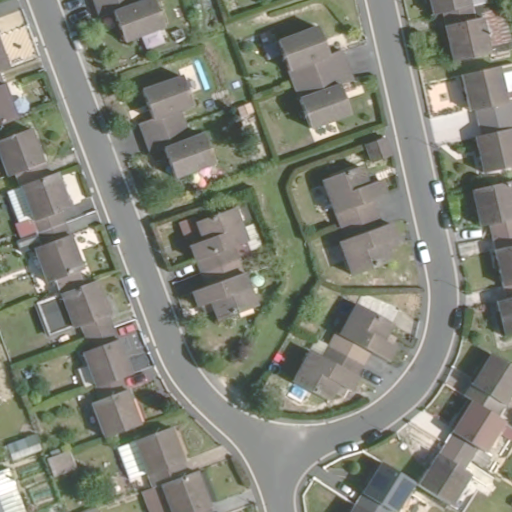}}\\
                        &Vegetation,\newline Agriculture\vspace{4mm}&40-60\vspace{4mm}&  \\
    \hline
      \multirow{2}{30mm}{\newline Sparse Residential}&Buildings&\vspace{2mm}10-30&\vspace{4mm}\multirow{2}{35mm}{\includegraphics[width=15mm, height=15mm]{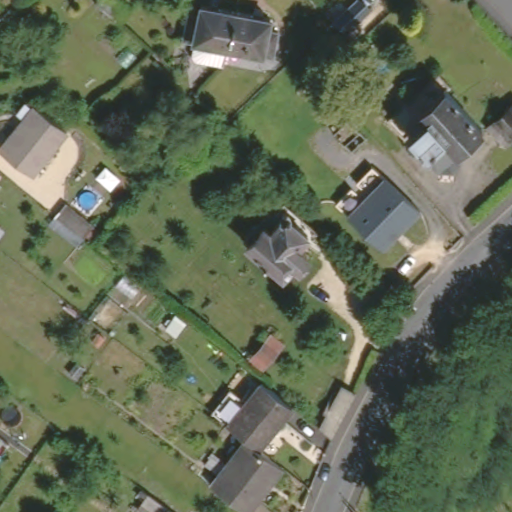} \includegraphics[width=15mm, height=15mm]{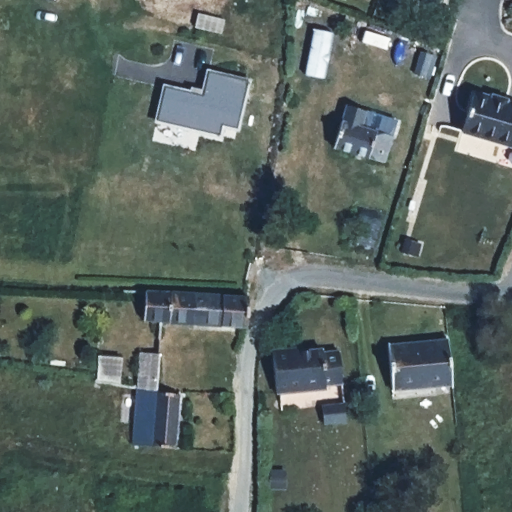}}\\
                        &Vegetation,\newline Agriculture\vspace{4mm}&70-90\vspace{4mm}& \\
    \hline
    \end{tabular}
    \caption{Concept dataset composition based on the land cover segmentation maps provided in the FLAIR dataset \cite{garioud2024flair}.}
    \label{tab:my_label}
\end{table*}

\section{Conceptual Sensitivities}
\label{sec:sensitivities}
\begin{figure*}[h]
    \centering    
    {\includegraphics[width=\textwidth]{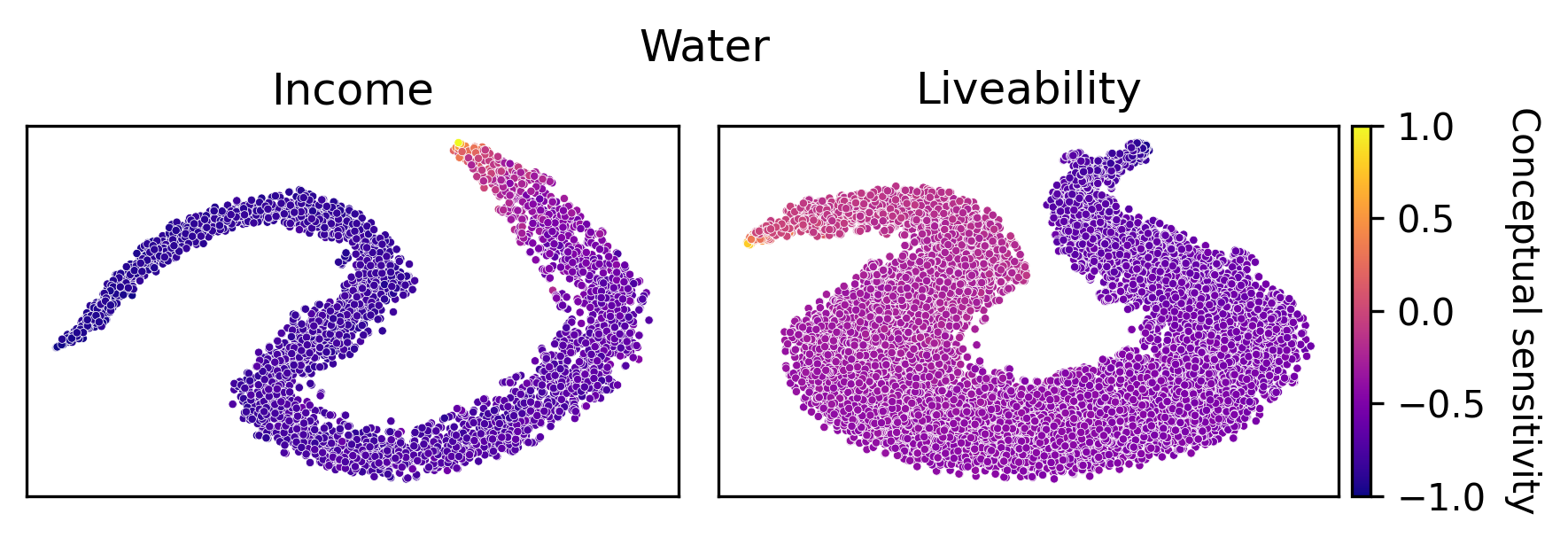}}
    \caption{\textbf{The TCAV sensitivity of the water concept for the income (left) and liveability (right) datasets.} The magnitude values are normalized in the range [-1, 1] by applying separate min-max normalization to the negative and the positive TCAV values, respectively.} 
    \label{fig:tcav_values_water}
\end{figure*}

\begin{figure*}[t]
    \centering    
    {\includegraphics[width=\textwidth]{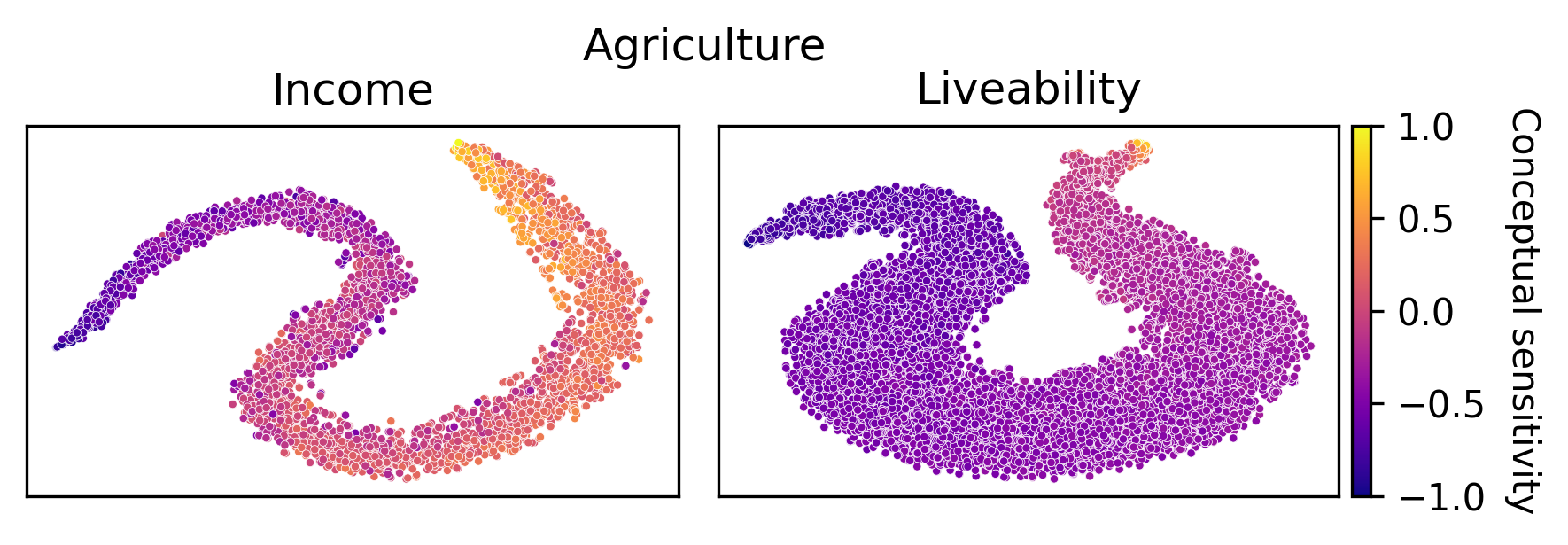}\label{fig:liveability_tcav_agriculture}}
    \caption{\textbf{The TCAV sensitivity of the agriculture concept for the income (left) and liveability (right) datasets.} The magnitude values are normalized in the range [-1, 1] by applying separate min-max normalization to the negative and the positive TCAV values, respectively.} 
    \label{fig:tcav_values_agriculture}
\end{figure*}

\begin{figure*}[t]
    \centering    
    {\includegraphics[width=\textwidth]{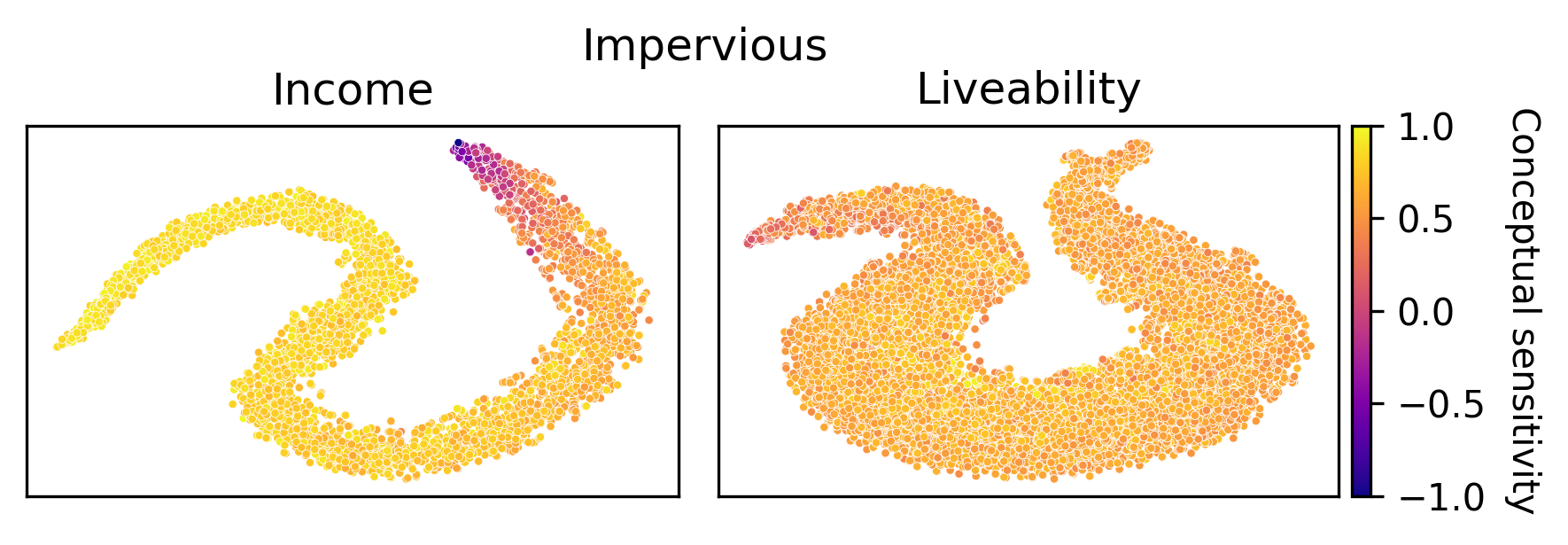}}
    \caption{\textbf{The TCAV sensitivity of the impervious surface concept for the income (left) and liveability (right) datasets.} The magnitude values are normalized in the range [-1, 1] by applying separate min-max normalization to the negative and the positive TCAV values, respectively.} 
    \label{fig:tcav_values_impervious}
\end{figure*}

\begin{figure*}[t]
    \centering    
    {\includegraphics[width=\textwidth]{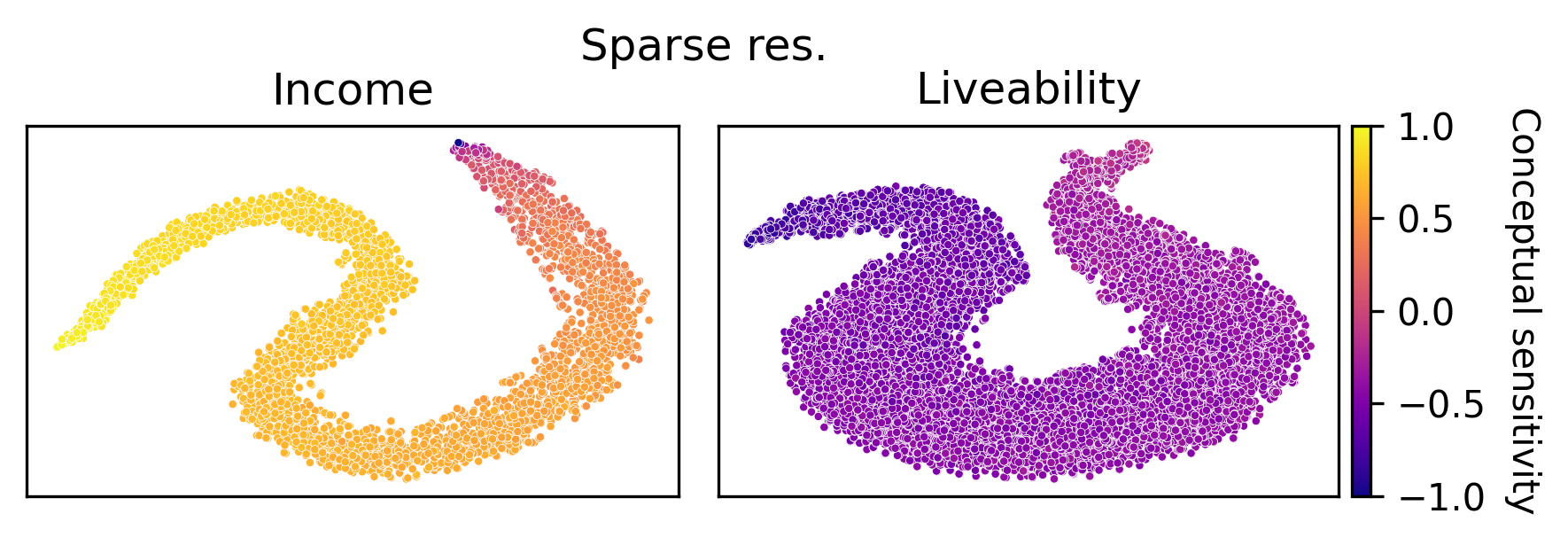}}
    \caption{\textbf{The TCAV sensitivity of the sparse residential concept for the income (left) and liveability (right) datasets.} The magnitude values are normalized in the range [-1, 1] by applying separate min-max normalization to the negative and the positive TCAV values, respectively.} 
    \label{fig:tcav_values_sparse_res}
\end{figure*}

\begin{figure*}[t]
    \centering    
    {\includegraphics[width=\textwidth]{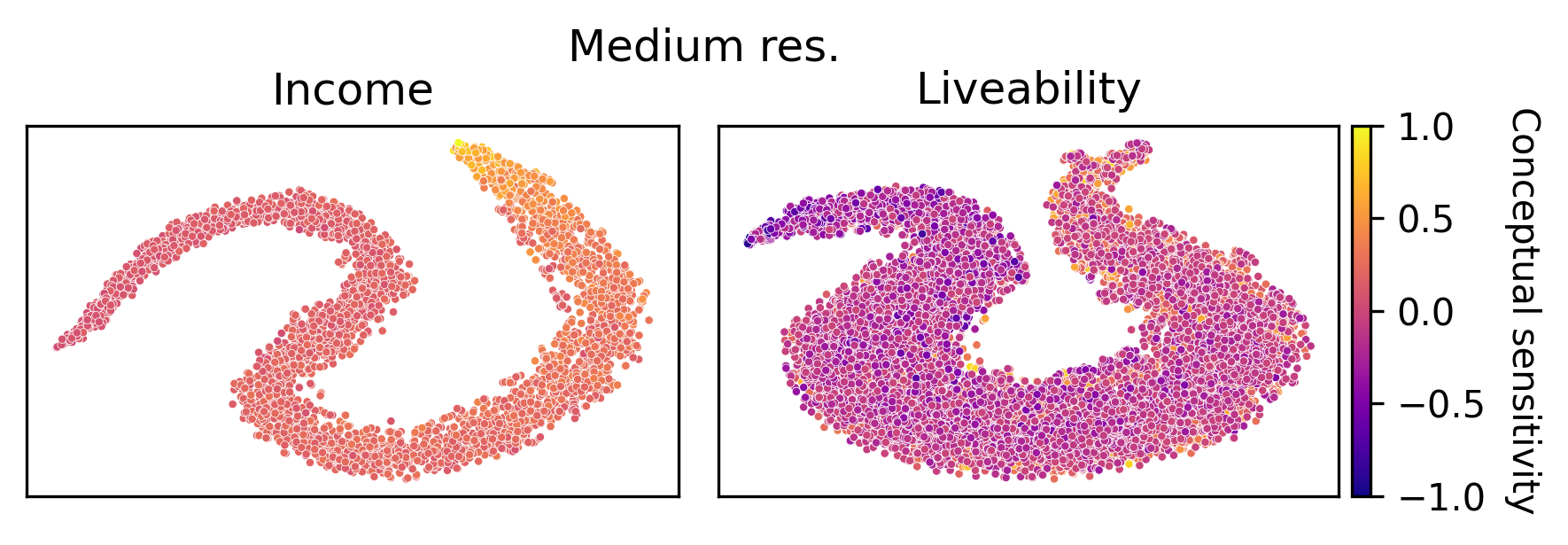}}
    \caption{\textbf{The TCAV sensitivity of the medium residential concept for the income (left) and liveability (right) datasets.} The magnitude values are normalized in the range [-1, 1] by applying separate min-max normalization to the negative and the positive TCAV values, respectively.} 
    \label{fig:tcav_values_medium_res}
\end{figure*}

\begin{figure*}[t]
    \centering    
    {\includegraphics[width=\textwidth]{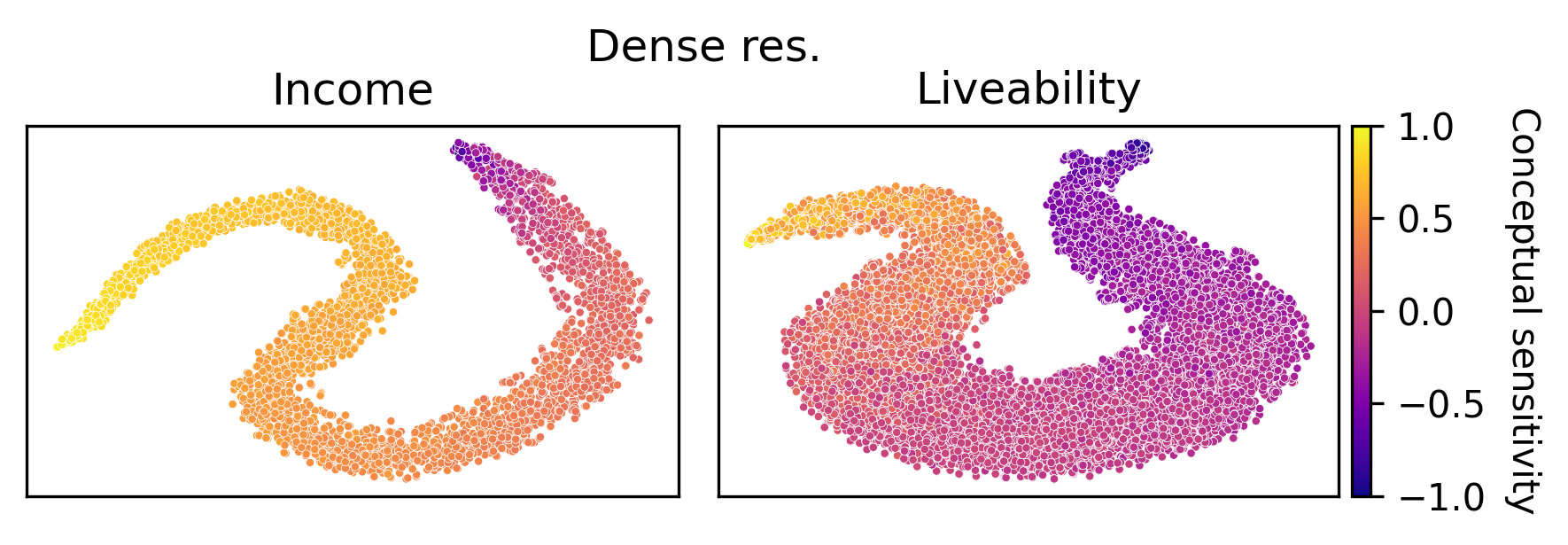}}
    \caption{\textbf{The TCAV sensitivity of the dense residential concept for the income (left) and liveability (right) datasets.} The magnitude values are normalized in the range [-1, 1] by applying separate min-max normalization to the negative and the positive TCAV values, respectively.} 
    \label{fig:tcav_values_dense_res}
\end{figure*}

Figures \ref{fig:tcav_values_water}, \ref{fig:tcav_values_agriculture}, \ref{fig:tcav_values_impervious}, \ref{fig:tcav_values_sparse_res}, \ref{fig:tcav_values_medium_res} and \ref{fig:tcav_values_dense_res} visualize the conceptual sensitivities of the Rank-N-Contrast pre-trained encoder in both datasets for the concepts of water, agriculture, impervious surfaces, sparse residential, medium residential and dense residential, respectively. 

\end{document}